\documentclass[sigconf]{acmart}

\def\BibTeX{{\rm B\kern-.05em{\sc i\kern-.025em b}\kern-.08emT\kern-.1667em\lower.7ex\hbox{E}\kern-.125emX}}
    
\copyrightyear{2019}
\acmYear{2019}
\setcopyright{acmlicensed}
\acmConference[WSDM '19]{WSDM '19}{February 11--15, 2019}{Melbourne, Australia}
\acmBooktitle{WSDM '19, February 11--15, 2019, Melbourne, Australia}
\acmPrice{15.00}
\acmDOI{10.1145/1122445.1122456}
\acmISBN{978-1-4503-9999-9/18/06}

\usepackage{multirow}

\hypersetup{draft}
\begin{document}

%
\title{Fake News Detection as Natural Language Inference}

%
\author{Kai-Chou Yang}
\affiliation{%
  \institution{National Cheng Kung University}
  \city{Tainan}
  \country{Taiwan}
}
\email{zake7749@gmail.com}

\author{Timothy Niven}
\affiliation{%
  \institution{National Cheng Kung University}
  \city{Tainan}
  \country{Taiwan}
}
\email{tim.niven.public@gmail.com}

\author{Hung-Yu Kao}
\affiliation{%
  \institution{National Cheng Kung University}
  \city{Tainan}
  \country{Taiwan}
}
\email{hykao@mail.ncku.edu.tw}

%
\renewcommand{\shortauthors}{Yang et al.}

%
\begin{abstract}
This report describes the entry by the Intelligent Knowledge Management (IKM) Lab in the WSDM 2019 Fake News Classification challenge. We treat the task as natural language inference (NLI). We individually train a number of the strongest NLI models as well as BERT. We ensemble these results and retrain with noisy labels in two stages. We analyze transitivity relations in the train and test sets and determine a set of test cases that can be reliably classified on this basis. The remainder of test cases are classified by our ensemble. Our entry achieves test set accuracy of 88.063\% for 3rd place in the competition.
\end{abstract}

%
%

%
\keywords{fake news, natural language inference, natural language processing}

%
\begin{teaserfigure}
  \centering
  \includegraphics[width=0.8\textwidth]{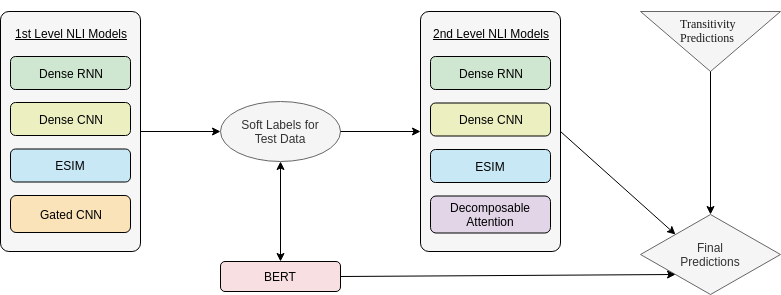}
  \caption{Overview of our method. High performing NLI models are independently trained and ensembled with a fine-tuned BERT model to determine soft labels, which are then used to fine-tune the original NLI models, BERT, and the Decomposable Attention model. These are then ensembled and combined with predictions made via observing transitivity relations.}
  \Description{Overview of our method}
  \label{fig:teaser}
\end{teaserfigure}

%
\maketitle

%
%

\section{Introduction}
The WSDM 2019 Fake News Classification challenge presents pairs of sentences requiring three-class prediction. The first sentence is the title of an article already known to be fake news. The second sentence is the title of another article, and the task is to decide whether it agrees with the original fake news, disagrees with it, or is unrelated. Sentences are in Mandarin and are drawn from news sources in China. English translations are provided, however they are noisy machine translations and we did not obtain good results using them. For this reason, we ignored the English sentences and just used the Mandarin.\footnote{Examples given in this report are in English and come from our own translations.}

This task can be viewed as a natural language inference task \cite{BowmanAPM15}, where the first sentence corresponds to the ``premise'' ($\mathbf{P}$) and the second to the ``hypothesis'' ($\mathbf{H}$). An example of a contradiction (disagreed) case taken from the training data:

\begin{tabular}{p{0.03\textwidth}p{0.37\textwidth}}
  $\mathbf{P}$ & Over 1,000 foreigners dissect children and steal their organs \\
  $\mathbf{H}$ & The rumour of 1,000 foreigners harvesting children's organs strikes again \\
\end{tabular}

Based on this observation we focused on the best performing neural models for SNLI\footnote{\hyperlink{https://nlp.stanford.edu/projects/snli/}{https://nlp.stanford.edu/projects/snli/}} as well as BERT \cite{DevlinCLT18}. We used these models in an ensemble and fine-tuned with soft labels in a process summarized in figure 1, that will be outlined in the following section.

We additionally explored transitive relations in the train and test datasets and found that classification on this basis can be used to reliably increase our accuracy as compared to using our final classifier's predictions.

%
%

\section{Method}

Our overall method is summarized in figure 1 and is explained in detail in this section, which aims to provide step-by-step instructions to accompany our published code\footnote{\hyperlink{https://github.com/zake7749/WSDM-Cup-2019}{https://github.com/zake7749/WSDM-Cup-2019}} for reproducing of our results.

In general we optimize with Adam \cite{KingmaB14}, use early stopping conditioned on validation set accuracy, and dropout \cite{SrivastavaHKS14a} for regularization. As we have a large number of models and hyperparameter settings, we refer the reader to our published code for further details.

\subsection{Embeddings}

We prepare multiple kinds of embedding at both the word and character level. We train models on all types independently and ensemble the results. This approach is motivated by the fact that Chinese word segmentation is a hard problem often resulting in noise and many out-of-vocabulary tokens. Ensembling the different information captured by these embedding techniques is also expected to reduce bias and variance.

At the word level, we use Tencent \cite{SongSLZ18} word embedding and the SGNS version of Chinese word embeddings \cite{LiZHLLD18}. These word embeddings are trained in an unsupervised manner on a large corpus drawn from different sources and have acceptable word coverage for this task. At the character level, we train a skip-gram model \cite{MikolovCCD13}, CBOW model, and FastText model \cite{BojanowskiGJ2017} with window sizes $5$, $7$, and $3$, respectively. The training data for character embedding training is composed of the set of sentences in train and test datasets. These three embeddings are then concatenated to form the final character-level embedding.

\subsection{First Level NLI Models}

\begin{table*}[t]
  \label{tab:freq}
  \caption{Test set accuracy of for NLI models}
  \begin{tabular}{l|ccc|ccc}
    \toprule
    \multirow{2}{*}{\textbf{Model}} & \multicolumn{3}{c|}{\textbf{First Level}} & \multicolumn{3}{c}{\textbf{Fine-tuned}} \\
    &Tencent&SGNS&Character&Tencent&SGNS&Character\\
    \midrule
    Decomposable Attention &         &         &          & 0.86730 & 0.86721 &  \\
    Dense RNN         & 0.85529 & 0.85620 & 0.85170  & 0.87704 & 0.87248 & 0.86809 \\
    Dense CNN         & 0.85082 & 0.84803 & 0.85287  & 0.87479 & 0.87114 & 0.86302 \\
    ESIM                   & 0.85738 & 0.85622 & 0.86053  & 0.87788 & 0.87334 & 0.87436 \\
    Gated CNN              & 0.84711 &         &          &  &  &  \\
    \bottomrule
  \end{tabular}
\end{table*}

We independently train a number of relatively high performing NLI models:

\begin{itemize}
    \item Dense RNN
    \item Dense CNN
    \item ESIM \cite{ChenZLWJ16}
    \item Gated CNN \cite{DauphinFAG16}
    \item Decomposable Attention \cite{ParikhTDU16}
\end{itemize}

The dense NLI architecture is an abstract version of \cite{KimHKK18}, which densely concatenates the features from different levels and repeats the comparison loop several times. The dense CNN shares the same general architecture as the dense RNN (figure 2) but uses a CNN encoder instead. For ESIM, Gated CNN and decomposable attention, we slightly modified the architectures to make them fit well on this task. Due to space requirements, we refer the reader to corresponding researches for details about their architectures.

\begin{figure}[t]
  \centering
  \includegraphics[width=0.4\textwidth]{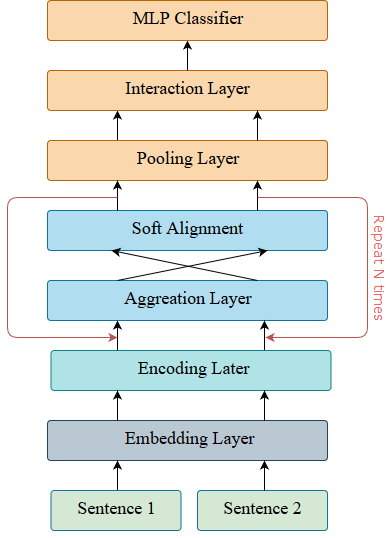}
  \caption{The general architecture of the Dense RNN and Dense CNN models.}
  \Description{Dense model architecture}
\end{figure}

\subsection{First Level Ensemble}

To ensemble the first level weak models, we use LightGBM\footnote{\hyperlink{https://github.com/Microsoft/LightGBM}{https://github.com/Microsoft/LightGBM}} and a densely connected feed-forward network. For each of our $10$ models, we concatenate the $3$ class label probabilities, yielding a $30$-dimensional input vector for each training and testing data point. For preventing data leakage, we choose the same validation set for early-stopping. The ensemble result achieves a test set accuracy of $86.741\%$.

\subsection{BERT}

We train the BERT base Chinese model\footnote{\hyperlink{https://github.com/huggingface/pytorch-pretrained-BERT}{https://github.com/huggingface/pytorch-pretrained-BERT}} for three epochs on the full training set. The learning rate was $5e$-$5$, maximum sequence length was $128$, and batch size was $32$. This achieved a test set accuracy of $86.689\%$. We did not perform an exhaustive search of hyperparameter space due to time constraints.

\subsection{Blending}

For the final output in the first level, we blend the predictions of (2.3) and (2.4). The blended predictions are the weighted sum of the BERT and ensemble predictions, with weights $0.42$ and $0.58$, respectively. We chose these weights by threshold search to make sure the blended result is an optimal mixture of both predictions. The test set accuracy is $86.963\%$.

\subsection{Second Level: Fine Tune NLI Models}

We took the predictions generated by step (2.5) as soft pseudo-labels and used them to fine-tune the pretrained NLI models in step (2.1). The training set was concatenated with the pseudo-labeled test set and the same validation set was used for early stopping. The fine-tuned results are reported in table 1. Note that the Decomposable attention model was added at this stage, initialized with random weights. Due to time constraints we did not train every combination of models and embeddings.

\subsection{Second Level Ensemble}

\begin{table}[t]
  \label{tab:freq2}
  \caption{Test set accuracy of for ensembles, BERT, blended models, and transitivity rules}
  \begin{tabular}{lc}
    \toprule
    \textbf{Model}&\textbf{Accuracy} \\
    \midrule
    Ensemble (1)               & 0.86741 \\
    BERT (1)                   & 0.86689 \\
    Blended (1)                & 0.86963 \\
    Ensemble (2)               & 0.87990 \\
    BERT (2)                   & 0.87484 \\
    Blended (2)                & 0.88019 \\
    Blended (2) + Transitivity & 0.88063 \\
    \bottomrule
  \end{tabular}
\end{table}

Once again, we used LightGBM and a multi-layer perceptron to perform ensembling with the output of the second level NLI models in the same manner as described in section (2.3). This achieved a test set accuracy of $87.990\%$.

\subsection{BERT Pseudo-Label Training}

We used the psueod-labels to continue fine tuning the BERT model from step (2.4). Whereas we tuned the NLI models on the pseudo-labeled test set only, BERT was trained on the whole concatenation of the entire training set and the pseudo-labeled test set, without any validation. We trained BERT for three epochs with the same hyperparameter settings as in step (2.4). This model achieved a test set score of $87.484\%$.

\subsection{Final Blending}

After fine-tuning BERT and the based NLI models, we once again blended their predictions to obtain our final predictions. The blending weights were $0.79$ and $0.21$, respectively. Just as in step (2.5) these weights were determined by threshold search. The blended result was $88.019\%$.

\subsection{Post-Processing: Transitive Relations}

We investigated transitivity relations in the data and found they were reliable enough to use as test set predictions. Figure 3 demonstrates the two types of relation we considered. For positive relations, we observed that if sentence $A$ agrees with sentence $B$, and $B$ agrees with sentence $C$, then $A$ should also agree with $C$. In the negative case, if $A$ disagrees with $B$, and $B$ agrees with $C$, the $A$ should also disagree with $C$.

We found that the positive case held for $99.9\%$ of the training data. The negative case held $99.7\%$ of the time. As there are sentence overlaps between the train and test sets, we are able to apply these rules recursively to generate predictions for $6,888$ data points in the test set. We expected these should be more reliable than our trained classifiers. We made a late submission with these predictions, labeling all other test set samples with a ``fake'' prediction label so only the samples labeled by transitivity rules were considered by the scorer. This submission achieved $93\%$ accuracy, a level of performance dramatically higher than any trained classifier in the competition, validating our expectation. Overall, we observed a $0.04\%$ increase in our test set accuracy using this method as compared to using our best classifier for all test set data points. The final accuracy was $88.063\%$.

\subsection{Results}

The results from all levels are summarized in table 2. We see a consistent improvement from successive ensembling and pseudo-label fine tuning. We note again that, due to time constraints we were unable to fully exploit all models we identified, nor conduct a suitably thorough hyperparameter search. We therefore expect our method could easily improve upon the results presented here.

\begin{figure}[t]
  \centering
  \includegraphics[width=0.4\textwidth]{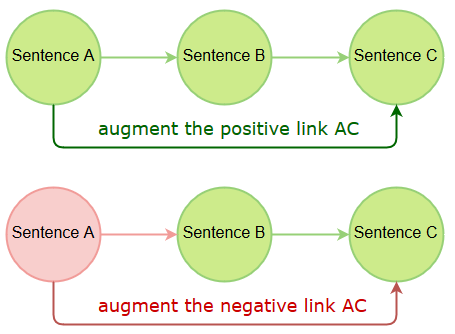}
  \caption{Positive and negative transitivity relations in the labeled data.}
  \Description{Dense model architecture}
\end{figure}

%
%

\section{Conclusion}

NLI models appear to be generally effective for this task. Ensemble via gradient boosting and fine-tuning with noisy labels proved to be very beneficial. However, due to time constraints we were unable to cover all the combinations we targeted and expect we could improve further. As the train and test sets have overlapping sentences, we were able to exploit transitive relations between them to reliably improve our performance.

For future work we intend to further investigate the transitivity method for data augmentation. Initial investigation revealed we can create an additional $700,000$ sentence ``agreed'' pairs and $19,000$ ``disagreed''. That is $7$ and $2$ times the number of original training data points, respectively.

%
\bibliographystyle{ACM-Reference-Format}
\bibliography{sample-base}


\begin{thebibliography}{12}


\ifx \showCODEN    \undefined \def \showCODEN     #1{\unskip}     \fi
\ifx \showDOI      \undefined \def \showDOI       #1{#1}\fi
\ifx \showISBNx    \undefined \def \showISBNx     #1{\unskip}     \fi
\ifx \showISBNxiii \undefined \def \showISBNxiii  #1{\unskip}     \fi
\ifx \showISSN     \undefined \def \showISSN      #1{\unskip}     \fi
\ifx \showLCCN     \undefined \def \showLCCN      #1{\unskip}     \fi
\ifx \shownote     \undefined \def \shownote      #1{#1}          \fi
\ifx \showarticletitle \undefined \def \showarticletitle #1{#1}   \fi
\ifx \showURL      \undefined \def \showURL       {\relax}        \fi
\providecommand\bibfield[2]{#2}
\providecommand\bibinfo[2]{#2}
\providecommand\natexlab[1]{#1}
\providecommand\showeprint[2][]{arXiv:#2}

\bibitem[\protect\citeauthoryear{Bojanowski, Grave, Joulin, and
  Mikolov}{Bojanowski et~al\mbox{.}}{2017}]%
        {BojanowskiGJ2017}
\bibfield{author}{\bibinfo{person}{Piotr Bojanowski}, \bibinfo{person}{Edouard
  Grave}, \bibinfo{person}{Armand Joulin}, {and} \bibinfo{person}{Tomas
  Mikolov}.} \bibinfo{year}{2017}\natexlab{}.
\newblock \showarticletitle{Enriching Word Vectors with Subword Information}.
\newblock \bibinfo{journal}{\emph{Transactions of the Association for
  Computational Linguistics}}  \bibinfo{volume}{5} (\bibinfo{year}{2017}),
  \bibinfo{pages}{135--146}.
\newblock
\showISSN{2307-387X}


\bibitem[\protect\citeauthoryear{Bowman, Angeli, Potts, and Manning}{Bowman
  et~al\mbox{.}}{2015}]%
        {BowmanAPM15}
\bibfield{author}{\bibinfo{person}{Samuel~R. Bowman}, \bibinfo{person}{Gabor
  Angeli}, \bibinfo{person}{Christopher Potts}, {and}
  \bibinfo{person}{Christopher~D. Manning}.} \bibinfo{year}{2015}\natexlab{}.
\newblock \showarticletitle{A large annotated corpus for learning natural
  language inference}.
\newblock \bibinfo{journal}{\emph{CoRR}}  \bibinfo{volume}{abs/1508.05326}
  (\bibinfo{year}{2015}).
\newblock
\showeprint[arxiv]{1508.05326}
\urldef\tempurl%
\url{http://arxiv.org/abs/1508.05326}
\showURL{%
\tempurl}


\bibitem[\protect\citeauthoryear{Chen, Zhu, Ling, Wei, and Jiang}{Chen
  et~al\mbox{.}}{2016}]%
        {ChenZLWJ16}
\bibfield{author}{\bibinfo{person}{Qian Chen}, \bibinfo{person}{Xiaodan Zhu},
  \bibinfo{person}{Zhen{-}Hua Ling}, \bibinfo{person}{Si Wei}, {and}
  \bibinfo{person}{Hui Jiang}.} \bibinfo{year}{2016}\natexlab{}.
\newblock \showarticletitle{Enhancing and Combining Sequential and Tree {LSTM}
  for Natural Language Inference}.
\newblock \bibinfo{journal}{\emph{CoRR}}  \bibinfo{volume}{abs/1609.06038}
  (\bibinfo{year}{2016}).
\newblock
\showeprint[arxiv]{1609.06038}
\urldef\tempurl%
\url{http://arxiv.org/abs/1609.06038}
\showURL{%
\tempurl}


\bibitem[\protect\citeauthoryear{Dauphin, Fan, Auli, and Grangier}{Dauphin
  et~al\mbox{.}}{2016}]%
        {DauphinFAG16}
\bibfield{author}{\bibinfo{person}{Yann~N. Dauphin}, \bibinfo{person}{Angela
  Fan}, \bibinfo{person}{Michael Auli}, {and} \bibinfo{person}{David
  Grangier}.} \bibinfo{year}{2016}\natexlab{}.
\newblock \showarticletitle{Language Modeling with Gated Convolutional
  Networks}.
\newblock \bibinfo{journal}{\emph{CoRR}}  \bibinfo{volume}{abs/1612.08083}
  (\bibinfo{year}{2016}).
\newblock
\showeprint[arxiv]{1612.08083}
\urldef\tempurl%
\url{http://arxiv.org/abs/1612.08083}
\showURL{%
\tempurl}


\bibitem[\protect\citeauthoryear{Devlin, Chang, Lee, and Toutanova}{Devlin
  et~al\mbox{.}}{2018}]%
        {DevlinCLT18}
\bibfield{author}{\bibinfo{person}{Jacob Devlin}, \bibinfo{person}{Ming{-}Wei
  Chang}, \bibinfo{person}{Kenton Lee}, {and} \bibinfo{person}{Kristina
  Toutanova}.} \bibinfo{year}{2018}\natexlab{}.
\newblock \showarticletitle{{BERT:} Pre-training of Deep Bidirectional
  Transformers for Language Understanding}.
\newblock \bibinfo{journal}{\emph{CoRR}}  \bibinfo{volume}{abs/1810.04805}
  (\bibinfo{year}{2018}).
\newblock
\showeprint[arxiv]{1810.04805}
\urldef\tempurl%
\url{http://arxiv.org/abs/1810.04805}
\showURL{%
\tempurl}


\bibitem[\protect\citeauthoryear{Kim, Hong, Kang, and Kwak}{Kim
  et~al\mbox{.}}{2018}]%
        {KimHKK18}
\bibfield{author}{\bibinfo{person}{Seonhoon Kim}, \bibinfo{person}{Jin{-}Hyuk
  Hong}, \bibinfo{person}{Inho Kang}, {and} \bibinfo{person}{Nojun Kwak}.}
  \bibinfo{year}{2018}\natexlab{}.
\newblock \showarticletitle{Semantic Sentence Matching with Densely-connected
  Recurrent and Co-attentive Information}.
\newblock \bibinfo{journal}{\emph{CoRR}}  \bibinfo{volume}{abs/1805.11360}
  (\bibinfo{year}{2018}).
\newblock
\showeprint[arxiv]{1805.11360}
\urldef\tempurl%
\url{http://arxiv.org/abs/1805.11360}
\showURL{%
\tempurl}


\bibitem[\protect\citeauthoryear{Kingma and Ba}{Kingma and Ba}{2014}]%
        {KingmaB14}
\bibfield{author}{\bibinfo{person}{Diederik~P. Kingma} {and}
  \bibinfo{person}{Jimmy Ba}.} \bibinfo{year}{2014}\natexlab{}.
\newblock \showarticletitle{Adam: {A} Method for Stochastic Optimization}.
\newblock \bibinfo{journal}{\emph{CoRR}}  \bibinfo{volume}{abs/1412.6980}
  (\bibinfo{year}{2014}).
\newblock
\showeprint[arxiv]{1412.6980}
\urldef\tempurl%
\url{http://arxiv.org/abs/1412.6980}
\showURL{%
\tempurl}


\bibitem[\protect\citeauthoryear{Li, Zhao, Hu, Li, Liu, and Du}{Li
  et~al\mbox{.}}{2018}]%
        {LiZHLLD18}
\bibfield{author}{\bibinfo{person}{Shen Li}, \bibinfo{person}{Zhe Zhao},
  \bibinfo{person}{Renfen Hu}, \bibinfo{person}{Wensi Li}, \bibinfo{person}{Tao
  Liu}, {and} \bibinfo{person}{Xiaoyong Du}.} \bibinfo{year}{2018}\natexlab{}.
\newblock \showarticletitle{Analogical Reasoning on Chinese Morphological and
  Semantic Relations}. In \bibinfo{booktitle}{\emph{Proceedings of the 56th
  Annual Meeting of the Association for Computational Linguistics (Volume 2:
  Short Papers)}}. \bibinfo{publisher}{Association for Computational
  Linguistics}, \bibinfo{pages}{138--143}.
\newblock
\urldef\tempurl%
\url{http://aclweb.org/anthology/P18-2023}
\showURL{%
\tempurl}


\bibitem[\protect\citeauthoryear{Mikolov, Chen, Corrado, and Dean}{Mikolov
  et~al\mbox{.}}{2013}]%
        {MikolovCCD13}
\bibfield{author}{\bibinfo{person}{Tomas Mikolov}, \bibinfo{person}{Kai Chen},
  \bibinfo{person}{Greg Corrado}, {and} \bibinfo{person}{Jeffrey Dean}.}
  \bibinfo{year}{2013}\natexlab{}.
\newblock \showarticletitle{Efficient Estimation of Word Representations in
  Vector Space}.
\newblock \bibinfo{journal}{\emph{CoRR}}  \bibinfo{volume}{abs/1301.3781}
  (\bibinfo{year}{2013}).
\newblock
\showeprint[arxiv]{1301.3781}
\urldef\tempurl%
\url{http://arxiv.org/abs/1301.3781}
\showURL{%
\tempurl}


\bibitem[\protect\citeauthoryear{Parikh, T{\"{a}}ckstr{\"{o}}m, Das, and
  Uszkoreit}{Parikh et~al\mbox{.}}{2016}]%
        {ParikhTDU16}
\bibfield{author}{\bibinfo{person}{Ankur~P. Parikh}, \bibinfo{person}{Oscar
  T{\"{a}}ckstr{\"{o}}m}, \bibinfo{person}{Dipanjan Das}, {and}
  \bibinfo{person}{Jakob Uszkoreit}.} \bibinfo{year}{2016}\natexlab{}.
\newblock \showarticletitle{A Decomposable Attention Model for Natural Language
  Inference}.
\newblock \bibinfo{journal}{\emph{CoRR}}  \bibinfo{volume}{abs/1606.01933}
  (\bibinfo{year}{2016}).
\newblock
\showeprint[arxiv]{1606.01933}
\urldef\tempurl%
\url{http://arxiv.org/abs/1606.01933}
\showURL{%
\tempurl}


\bibitem[\protect\citeauthoryear{Song, Shi, Li, and Zhang}{Song
  et~al\mbox{.}}{2018}]%
        {SongSLZ18}
\bibfield{author}{\bibinfo{person}{Yan Song}, \bibinfo{person}{Shuming Shi},
  \bibinfo{person}{Jing Li}, {and} \bibinfo{person}{Haisong Zhang}.}
  \bibinfo{year}{2018}\natexlab{}.
\newblock \showarticletitle{Directional Skip-Gram: Explicitly Distinguishing
  Left and Right Context for Word Embeddings}. In
  \bibinfo{booktitle}{\emph{Proceedings of the 2018 Conference of the North
  American Chapter of the Association for Computational Linguistics: Human
  Language Technologies, Volume 2 (Short Papers)}}.
  \bibinfo{publisher}{Association for Computational Linguistics},
  \bibinfo{pages}{175--180}.
\newblock
\urldef\tempurl%
\url{https://doi.org/10.18653/v1/N18-2028}
\showDOI{\tempurl}


\bibitem[\protect\citeauthoryear{Srivastava, Hinton, Krizhevsky, Sutskever, and
  Salakhutdinov}{Srivastava et~al\mbox{.}}{2014}]%
        {SrivastavaHKS14a}
\bibfield{author}{\bibinfo{person}{Nitish Srivastava},
  \bibinfo{person}{Geoffrey Hinton}, \bibinfo{person}{Alex Krizhevsky},
  \bibinfo{person}{Ilya Sutskever}, {and} \bibinfo{person}{Ruslan
  Salakhutdinov}.} \bibinfo{year}{2014}\natexlab{}.
\newblock \showarticletitle{Dropout: A Simple Way to Prevent Neural Networks
  from Overfitting}.
\newblock \bibinfo{journal}{\emph{Journal of Machine Learning Research}}
  \bibinfo{volume}{15} (\bibinfo{year}{2014}), \bibinfo{pages}{1929--1958}.
\newblock
\urldef\tempurl%
\url{http://jmlr.org/papers/v15/srivastava14a.html}
\showURL{%
\tempurl}


\end{thebibliography}

\end{document}